\documentclass[11pt,a4paper]{article}
\usepackage[hyperref]{conll2018}
\usepackage{times}
\usepackage{latexsym}
\usepackage{url}

\usepackage[T1]{fontenc}
\usepackage[utf8]{inputenc}
\usepackage{times}
\usepackage{latexsym}
\usepackage{url}
\usepackage{amsmath, amsthm, amssymb, bbm, physics}
\usepackage{algorithm}
\usepackage{algpseudocode}

\usepackage{graphicx, subcaption}
\usepackage{array, booktabs, multirow, makecell}
\usepackage{enumitem}
\usepackage{nameref}  
\usepackage{xspace}  
\usepackage{longtable} 
\usepackage{colortbl}
\graphicspath{{figures/}}

\usepackage[normalem]{ulem} 
\usepackage{color}\definecolor{RED}{rgb}{1,0,0}\definecolor{BLUE}{rgb}{0,0,1}

\makeatletter
\DeclareRobustCommand\onedot{\futurelet\@let@token\@onedot}
\def\@onedot{\ifx\@let@token.\else.\null\fi\xspace}

\def\eg{\emph{e.g}\onedot} 
\def\ie{\emph{i.e}\onedot}

\makeatother

\def\bea{\begin{eqnarray}}
\def\eea{\end{eqnarray}}

\def\fig#1{Figure~\ref{fig:#1}}
\def\tab#1{Table~\ref{tab:#1}}
\def\sect#1{Section~\ref{sec:#1}}

\def\Eq#1{Eq.~(\ref{eq:#1})}

\newcommand{\src}{\boldsymbol s}
\newcommand{\trg}{\boldsymbol t}
\newcommand{\lsrc}{{|\boldsymbol s|}}
\newcommand{\ltrg}{{|\boldsymbol t|}}

\newcommand{\V}{\mathcal V}
\newcommand{\R}{\mathbb{R}}

\newcolumntype{H}{>{\setbox0=\hbox\bgroup}c<{\egroup}@{}}

\usepackage{tikz}
\usetikzlibrary{shapes,arrows,fit,calc,positioning,automata}
\tikzstyle{neuron}=[draw, circle, fill=white, inner sep=5,outer sep=0, align=center]
\tikzstyle{pointer}=[-stealth]
\tikzstyle{state}=[draw, rectangle, fill=white, inner sep=2,outer sep=0, align=center]
\tikzstyle{RES}=[draw,circle,append after command={
    [shorten >=\pgflinewidth, shorten <=\pgflinewidth,]
    (\tikzlastnode.north) edge (\tikzlastnode.south)
    (\tikzlastnode.east) edge (\tikzlastnode.west)
}
]
\tikzstyle{DOT}=[draw,circle, inner sep=2pt,append after command={
    [shorten >=\pgflinewidth, shorten <=\pgflinewidth,]
    (\tikzlastnode.north east) edge (\tikzlastnode.south west)
    (\tikzlastnode.north west) edge (\tikzlastnode.south east)
}
]

\aclfinalcopy

\title{Pervasive Attention: 2D Convolutional Neural Networks\\
for Sequence-to-Sequence Prediction} 

\newcommand{\ours}{{Pervasive Attention}~}

\author{Maha Elbayad\textsuperscript{1,2} \hspace{20pt}
        Laurent Besacier\textsuperscript{1} \hspace{20pt}
        Jakob Verbeek\textsuperscript{2} \\
        Univ.\ Grenoble Alpes, CNRS, Grenoble INP, Inria, LIG, LJK, F-38000 Grenoble France\\
        \textsuperscript{1} \tt{firstname.lastname@univ-grenoble-alpes.fr}\\
        \textsuperscript{2} \tt{firstname.lastname@inria.fr}
         }
\date{}

\begin{document}
\maketitle

\begin{abstract}
Current state-of-the-art machine translation systems are based on encoder-decoder architectures, that first encode the input sequence, and then generate an output sequence based on the input encoding. Both are interfaced with an attention mechanism that recombines a fixed encoding of the source tokens based on the decoder state. We propose an alternative approach which instead relies on a single 2D convolutional neural network across both sequences. Each layer of our network re-codes source tokens on the basis of the output sequence produced so far. Attention-like properties are therefore pervasive throughout the network. Our model yields results that are competitive with state-of-the-art encoder-decoder systems, while being conceptually simpler and having fewer parameters.
\end{abstract}

\section{Introduction}
\label{sec:intro}
Deep neural networks have made a profound impact on natural language processing technology in general, and machine translation in particular \cite{kalchbrenner13acl,sutskever14nips,cho14emnlp,jean15acl,lecun15nature}.  
Machine translation (MT) can be seen as a sequence-to-sequence prediction problem, where the source and target sequences are of different and variable length.
Current state-of-the-art approaches are based on encoder-decoder architectures~\cite{kalchbrenner13acl,sutskever14nips,cho14emnlp,bahdanau15iclr}.  
The encoder ``reads'' the variable-length source sequence and maps it into a vector representation.  The decoder
takes this vector as input and ``writes'' the target sequence,
updating 
its state each step with the most recent word that it generated.
The basic encoder-decoder model is generally equipped with an attention model \cite{bahdanau15iclr}, which repetitively re-accesses the source sequence during the decoding process.
Given the current state of the decoder, a probability distribution over the elements in the source sequence is computed, which is then used to select or aggregate features of these elements into a single ``context'' vector that is used by the decoder.
Rather than relying on the global representation of the source sequence, the attention mechanism allows the decoder to ``look back'' into the source sequence and focus on salient positions. Besides this inductive bias, the attention mechanism bypasses the problem of vanishing gradients that most recurrent architectures encounter.

However, the current attention mechanisms have limited modeling abilities and are generally a simple weighted sum of the source representations \cite{ bahdanau15iclr, luong15emnlp}, where the weights are the result of a shallow matching between source and target elements. The attention module re-combines the same source token codes and is unable to re-encode or re-interpret the source sequence while decoding.
\begin{figure}
\begin{center}
\includegraphics[width=\linewidth]{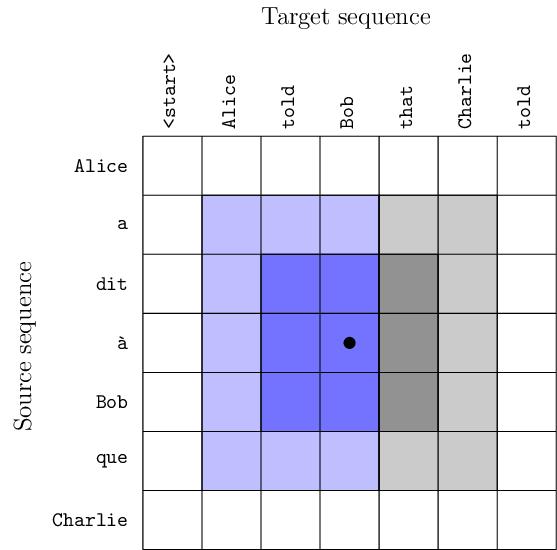}
\end{center}
\caption{Convolutional layers in our model use masked $3\!\times\!3$ filters so that features are only computed from previous output symbols. 
Illustration of the receptive fields after one (dark blue) and two layers (light blue), together with the masked part of the field of view of a normal $3\!\times\!3$ filter (gray).}
\label{fig:teaser}
\end{figure}

To address these limitations, we propose an alternative neural MT architecture, based on deep 2D convolutional neural networks (CNNs). 
The product space of the positions in source and target sequences defines the 2D grid over which the network is defined. 
The convolutional filters are masked to prohibit accessing information derived from future tokens in the target sequence, obtaining an autoregressive model akin to generative models for images and audio waveforms \cite{oord16ssw,oord16icml}.
See \fig{teaser} for an illustration.

This approach allows us to learn deep feature hierarchies based on a stack of 2D convolutional layers, and benefit from parallel computation during training.
Every layer of our network computes features of the the source tokens, based on the target sequence produced so far, and uses these to predict the next output token. 
Our model therefore has attention-like capabilities by construction, that are pervasive throughout the layers of the network, rather than using an ``add-on'' attention model.

We validate our model with experiments on the IWSLT 2014 German-to-English (De-En) and English-to-German(En-De) tasks.
We improve on state-of-the-art encoder-decoder models with attention, while being conceptually simpler and having fewer parameters.

In the next section we will discuss related work, before presenting our approach in detail in \sect{model}.
We present our experimental evaluation results in \sect{experiments}, and conclude in \sect{conclusion}.

\section{Related work}

The predominant neural architectures in machine translation are recurrent encoder-decoder networks \citep{graves12arxiv,sutskever14nips,cho14emnlp}.
The encoder is a recurrent neural network (RNN) based on gated recurrent units \citep{hochreiter97neco,cho14emnlp} to  map the input sequence into a vector representation. 
Often a bi-directional RNN \citep{schuster97sp} is used, which consists of two RNNs that process the input in opposite directions, and the final states of both RNNs are concatenated as the input encoding. 
The decoder consists of a second RNN, which takes the input encoding, and sequentially samples the output sequence one token at a time
whilst updating its state.

While best known for their use in visual recognition models, 
\citep{oord16ssw, salimans17iclr, reed17icml, oord16nips}.
Recent works also introduced convolutional networks to natural language processing. The first convolutional apporaches to encoding variable-length sequences consist of stacking word vectors, applying 1D convolutions then aggregating with a max-pooling operator over time~\cite{collobert08icml, kalchbrenner14acl, kim14acl}.
For sequence generation, the works of \citet{ranzato16iclr,bahdanau17iclr,gehring17acl} mix a convolutional encoder with an RNN decoder. The first entirely convolutional encoder-decoder models where introduced by \citet{kalchbrenner16arxiv}, but they did not  improve over state-of-the-art recurrent architectures. \citet{gehring17icml} outperformed deep LSTMs for machine translation 1D CNNs with gated linear units \citep{meng15acl, oord16nips, dauphin17icml} in both the encoder and decoder modules.

\begin{figure}
\centering
\includegraphics[width=\linewidth]{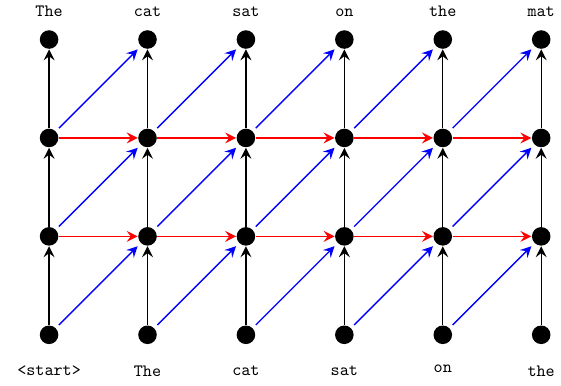}
\caption{Illustration of decoder network topology with two hidden layers, nodes at bottom and top represent input and output respectively.
Horizontal connections are used for RNNs, diagonal connections for convolutional networks. Vertical connections are used in both cases.
Parameters are shared across time-steps (horizontally), but not across layers (vertically).
\label{fig:autoRNN}}
\end{figure}

Such CNN-based models differ from their RNN-based counterparts in that temporal connections are placed between layers of the network, rather than within layers. See \fig{autoRNN} for a conceptual illustration.  This apparently small difference in connectivity has two important consequences. First, it makes the field of view grow linearly across layers in the convolutional network, while it is unbounded within layers in the recurrent network. Second, while the activations in the RNN can only be computed in a sequential manner, they can be computed in parallel across the temporal dimension in the convolutional case.

In all the recurrent or convolutional models mentioned above, each of the input and output sequences are processed separately as a one-dimensional sequence by the encoder and decoder respectively.
Attention mechanisms \citep{bahdanau15iclr, luong15emnlp, xu15icml} were introduced as an interface between the encoder and decoder modules. During encoding, the attention model finds which hidden states from the source code are the most salient for generating the next target token. This is achieved by evaluating a ``context vector'' which, in its most basic form, is a weighted average of the source features. The weights of the summation are predicted by a small neural network that scores these features conditioning on the current decoder state.

\citet{vaswani17nips} propose an architecture relying entirely on attention.
Positional input coding together with self-attention~\cite{parikh16emnlp, lin17iclr} replaces recurrent and convolutional layers. 
\citet{huang18iclr} use an attention-like gating mechanism to alleviate an assumption of monotonic alignment in the phrase-based translation model of~\citet{wang17icml}.
\citet{deng18arxiv} treat the sentence alignment as a latent variable which they infer using a variational inference network during training to optimize a variational lower-bound on the log-likelihood.

\paragraph{Beyond uni-dimensional encoding/decoding.}
The idea of building a 2D grid from parallel sequences (as in \fig{teaser}) is used in different NLP tasks especially for scoring parallel texts. This includes works on semantic matching, paraphrase identification and machine translation. ARC-II of \citet{hu14nips} has 1D convolutions applied to each sequence separately before a series of 2D convolutions and max-poolings are followed by an MLP to estimate the matching score. They interestingly highlighted the desirable property of letting the sequences `meet' before their representations mature.
\citet{he16naacl, wan16aaai} first encode the sequences with Bi-LSTMs then evaluate pairwise similarities between the words of the two sequences to build an interaction grid. While \citet{he16naacl} process the grid with a two-dimensional CNN, \citet{wan16aaai} directly use k-max pooling to aggregate and then score the pair. Similarly, for sequence alignment, \citet{levy17icml} use LSTM hidden states as tokens representations and, similar to our work, concatenate pairwise representations and feed their input grid to a 2D convolutional network followed by a soft-max to estimate soft-alignment probablities. Recently in question-answering, \citet{raison18arxiv} weaved two Bi-LSTMs, one along the context dimension and the other along the question dimension in order to identify a response span in the context.

More related to our work on machine translation, \citet{kalchbrenner16iclr} proposed the `reencoder' network where a Grid LSTM processes both sequences along its first and second dimension, allowing the model to re-encode the source sequence as it advances along the target dimension. They also observed that such a structure implements an implicit form of attention.
\citet{wu17arxiv} used a CNN over the 2D source-target representation, but only as a discriminator in an adversarial training setup. Similar to semantic matching models, they do not use masked convolutions, since their CNN is used to predict if a given source-target pair is a human or machine translation. 
Concurrently with our work, \citet{bahar18emnlp} used a 2DLSTM layer to jointly process the source and target sequences with a similar two-dimensional layout.

\section{Translation by 2D Convolution}
\label{sec:model}

In this section we present our 2D CNN translation model in detail.

\paragraph{Input source-target tensor.}
Given the source and target pair $(\src,\trg)$ of lengths $\lsrc$ and $\ltrg$ respectively, we first embed the tokens in $d_s$ and $d_t$ dimensional spaces via look-up tables. 
The word embeddings $\{x_1,\ldots, x_\lsrc\}$ and $\{y_1,\ldots,y_\ltrg\}$ are then concatenated to form a 3D tensor $X\in\R^{\ltrg\times\lsrc\times f_0}$, with $f_0=d_t+d_s$, where
\begin{align}
    X_{ij} = [y_i \;\; x_j].
\end{align}
This joint unigram encoding is the input to our convolutional network.

\paragraph{Convolutional layers.}

We use the DenseNet~\cite{huang17cvpr} convolutional architecture, which is the state of the art for image classification tasks.
Layers are densely connected, meaning that each layer takes as input the  activations of all the preceding layers, rather than just the last one, to produce its $g$ feature maps. The parameter $g$ is called the ``growth rate'' as it is the number of appended channels to the network's output at each layer.
The long-distance connections in the network improve gradient flow to early network layers during training, which is beneficial for deeper networks.

\begin{figure}
\begin{center}
\resizebox{\columnwidth}{!}{
\begin{tikzpicture}
    \node (in) [rectangle] {};
    \node (0) [right=5mm of in, neuron] {};
    \path[pointer] (in) edge node {} (0);
    \foreach \x in {0,...,3}{
     \pgfmathtruncatemacro{\next}{\x+1}
\pgfmathtruncatemacro{\prev}{\x+-1}
     \node (\next)[neuron, right=1cm of \x]  {};
     \draw [pointer] (\x) -- (\next);
    }
    \draw[pointer] (4.east) -- +(4mm,0) ;
   \foreach \x in {0,...,2}{
     \pgfmathtruncatemacro{\next}{\x+2}
     \foreach \y in {\next,...,4}{
            \draw[->, gray] (\x) to [out=-60,in=-120] (\y);
        }

}
\end{tikzpicture}
}

\resizebox{.75\columnwidth}{!}{
    \begin{tikzpicture}
    \node (in) [state] {\rotatebox{90}{Input}};
    \node (bn1) [right=3mm of in,  state] {\rotatebox{90}{BN}};
    \node (relu1) [right=3mm of bn1,  state] {\rotatebox{90}{ReLU}};
    \node (conv1) [right=3mm of relu1,  state] {\rotatebox{90}{Conv(1)}};
    \node (bn2) [right=3mm of conv1,  state] {\rotatebox{90}{BN}};
    \node (relu2) [right=3mm of bn2,  state] {\rotatebox{90}{ReLU}};
    \node (conv2) [right=3mm of relu2,  state] {\rotatebox{90}{Conv(k)}};
    \node (dp) [right=3mm of conv2,  state] {\rotatebox{90}{Dropout}};
    \node (out) [rectangle, right=2mm of dp] {};
    \path[->]
          (in) edge node {} (bn1)
		  (bn1) edge node {} (relu1)
          (relu1) edge node {} (conv1)
          (conv1) edge node {} (bn2)
          (bn2) edge node {} (relu2)
          (relu2) edge node {} (conv2)
          (conv2) edge node {} (dp)
          (dp) edge node {} (out);
    \end{tikzpicture}
}
\end{center}
\caption{Architecture of the DenseNet at block level (top), and within each block (bottom).
}
\label{fig:dense}
\end{figure}
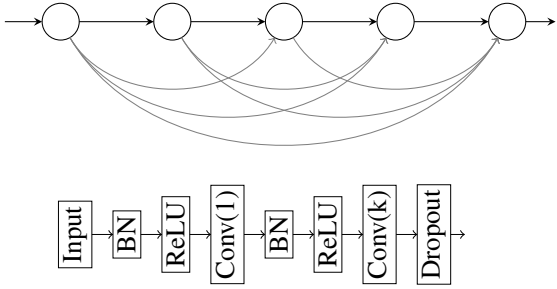

Each layer first batch-normalizes \cite{ioffe15icml} its input and apply a ReLU \cite{nair10icml} non-linearity. 
To reduce the computation cost, each layer first computes $4g$ channels using a $1\!\times\!1$ convolution from the $f_0 + (l-1)g$ input channels to layer $l\in\{1,\dots,L\}$. This is followed by a second batch-normalization and ReLU non-linearity. 
The second convolution has $(k\times \lceil \frac{k}{2}\rceil)$ kernels, \ie masked as illustrated in \fig{teaser}, and generates the $g$ output features maps to which we apply dropout \cite{srivastava14jmlr}.
The architecture of the densely connected network is illustrated in \fig{dense}.

We optionally use gated linear units \citep{dauphin17icml} in both convolutions, these double the number of output channels, and we use half of them to gate the other half.

\paragraph{Target sequence prediction.}
Starting from the initial $f_0$ feature maps, each layer $l\in\{1,\ldots,L\}$ of our DenseNet produces a tensor $H^l$ of size $\ltrg\!\times\!\lsrc\!\times\!f_l$, 
where $f_l$ is the number of output channels of that layer.
To compute a distribution over the tokens in the output vocabulary, 
we need to collapse the second dimension of the tensor, which is given by the variable length of the input sequence, to retrieve a unique encoding for each target position.

The simplest aggregation approach is to apply max-pooling over the input sequence to obtain a tensor $H^\text{pool}\in\R^{\ltrg\times f_L}$, \ie 
\begin{align}
    &H_{id}^\text{pool}= \max_{j\in\{1,\dots,\lsrc\}}H^L_{ijd}. \label{eq:maxpool}
\end{align}
Alternatively, we can use average-pooling over the input sequence:
\begin{align}
    &H_{id}^\text{pool}= \frac{1}{\sqrt{\lsrc}}\sum_{j\in\{1,\dots,\lsrc\}}H^L_{ijd}.
    \label{eq:avgpool}
\end{align}
The scaling with the inverse square-root of the source length acts as a variance stabilization term, which we find to be more effective in practice than a simple averaging.

The pooled features are then transformed to predictions over the output vocabulary $\V$, 
by linearly mapping them with a matrix $E\in\R^{|\V|\times f_L}$ to the vocabulary dimension $|\V|$, and then applying a soft-max. 
Thus the probability distribution over $\V$ for the $i$-th output token is obtained as 
\begin{align}
p_i = \text{SoftMax}(EH_i^\text{pool}).
\end{align}
Alternatively, we can use $E$ to project to dimension $d_t$, and then multiply with the target word embedding matrix used to define the input tensor. 
This reduces the number of parameters and generally improves the performance.

\paragraph{Implicit sentence alignment.} \label{para:align}
For a given output token position $i$, the max-pooling operator of \Eq{maxpool} partitions the $f_L$ channels 
by assigning them across the source tokens $j$.  
Let us define 
\[B_{ij}=\{ d\in \{1,\ldots, f_L\} |\: j=\arg\max(H^L_{ijd}) \}\]
as the channels assigned to source token $j$ for output token $i$.
The energy that enters into the soft-max to predict token $w\in\V$ for the $i$-th output position is given by 
\begin{eqnarray}
e_{iw} & = &  \sum_{d \in\{1,\dots,f_L\}}  E_{wd} H^\text{pool}_{id} \\
 & = & \sum_{j\in\{1,\dots,\lsrc\}} \sum_{d\in B_{ij}} E_{wd} H^L_{ijd}.
\end{eqnarray}
The total contribution of the $j$-th input token is thus given by 
\begin{align}
\alpha_{ij} = \sum_{d\in B_{ij}} E_{wd} H^L_{ijd},
\label{eq:implicit}
\end{align}
where we dropped the dependence on $w$ for simplicity. 
As we will show experimentally in the next section, visualizing the values $\alpha_{ij}$ for the ground-truth output tokens, we can recover an implicit sentence alignment used by the model.

\paragraph{Self attention.}
Besides  pooling we can collapse the source dimension of the feature tensor with an attention mechanism. 
This mechanism will generate a tensor $H^\textrm{att}$ that can be used instead of, or concatenated with, $H^\text{Pool}$.

We use the self-attention approach of \citet{lin17iclr}, 
which for output token $i$ computes the attention vector $\rho_i \in\R^\lsrc$ from the  activations $H^L_i$:
\begin{eqnarray}
\rho_i  & = &\text{SoftMax}\left(H_{i}^L  w  + b  \mathbbm 1_{\lsrc}\right),\label{eq:attention1}\\
H_i^\text{att}  & = & \sqrt{\lsrc} \rho_{i}^\top H_i^L,
\label{eq:attention}
\end{eqnarray}
where $w \in\R^{f_L}$ and $b\in\R$ are parameters of the attention mechanism.
Scaling of attention vectors with the square-root of the source length was also used by \citet{gehring17icml}, and we found it effective here as well  as in the average-pooling case.

\section{Experimental evaluation}
\label{sec:experiments}

In this section, we present our experimental setup, 
followed by quantitative results, qualitative examples of implicit sentence alignments from our model, and a comparison to the state of the art.

\subsection{Experimental setup}
\paragraph{Data and pre-processing.}

We experiment with the IWSLT 2014 bilingual dataset~\cite{cettolo14iwslt}, which contains transcripts of TED talks aligned at sentence level, and translate between German (De) and English (En) in both directions.
Following the setup of \cite{edunov18naacl}, sentences longer than 175 words and pairs with length ratio exceeding 1.5 were removed from the original data.
There are 160+7K training sentence pairs, 7K of which are separated and used for validation/development.
We report results on a test set of 6,578 pairs obtained by concatenating TED.dev2010, TEDX.dev2012 and TED.tst2010-2012.
We tokenized and lower-cased all data using the standard scripts from the Moses toolkit \citep{koehn07acl}.

For open-vocabulary translation, we segment sequences using byte pair encoding \cite{sennrich16acl} with 14K merge operations following two approaches. The first (V1), similar to \citet{edunov18naacl, deng18arxiv}, is a joint encoding i.e. applied to the concatenation of source and target texts. This results in a German and English vocabularies of around 12K and 8.8K types respectively. 
The second approach (V2) encodes each language independently resulting in a German and English vocabularies of 13.3K and 13.8K respectively.

\paragraph{Implementation details.}
Unless stated otherwise, we use DenseNets with masked convolutional filters of size $5\!\times\!3$, as given by the light blue area in \fig{teaser}.
To train our models for the ablation study, we use maximum likelihood estimation (MLE) with Adam $(\beta_1=0.9, \beta_2=0.999, \epsilon=1e^{-8})$ starting with a learning rate of $5e^{-4}$ that we scale by a factor of 0.8 if no improvement is noticed on the validation loss after three evaluations; we evaluate every 8K updates.
For faster training and due to the increased computational requirements, from $\mathcal O(|x|+|y|)$ of encoder-decoder models to $\mathcal O(|x|.|y|)$, we only read sequences up to 80 positions. We also downsample the initial grid channels by half to reduce the number of input channels to every dense block, thus requiring less memory.
After training all models for 40 epochs, the best performing model on the validation set is usd to decode with a beam-search of width 5. We measure translation quality using the BLEU metric \cite{papineni02acl}.
\begin{figure*}[!t]
\begin{center}
\begin{subfigure}{0.32\textwidth}
    \caption{\label{fig:ablation:emb}$L=20,\; g=32$ } 
\includegraphics[width=\textwidth]{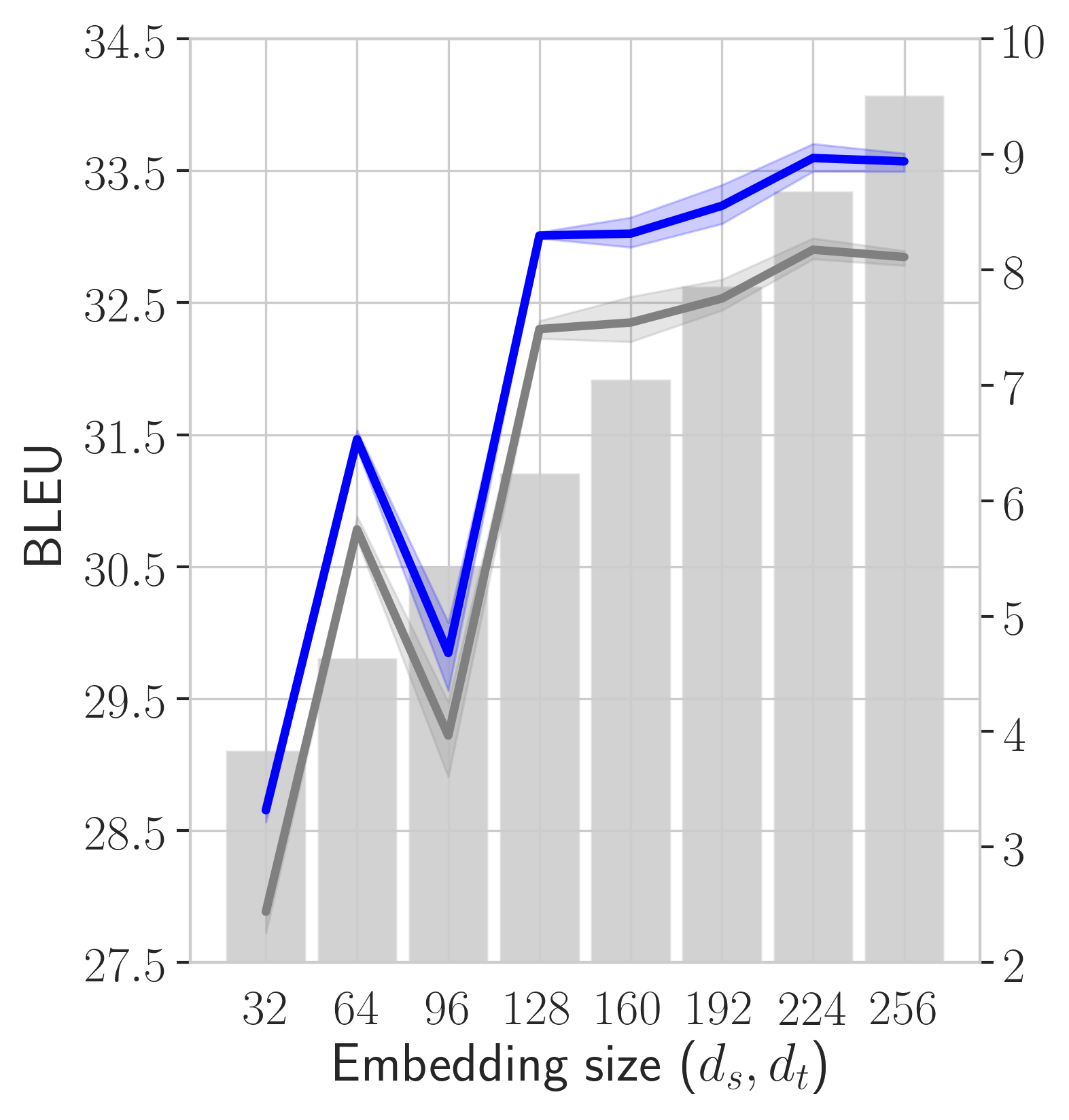}
\end{subfigure}
\begin{subfigure}{0.32\textwidth}
    \caption{\label{fig:ablation:growth}$L=20,\; d=128$}
\includegraphics[width=\textwidth]{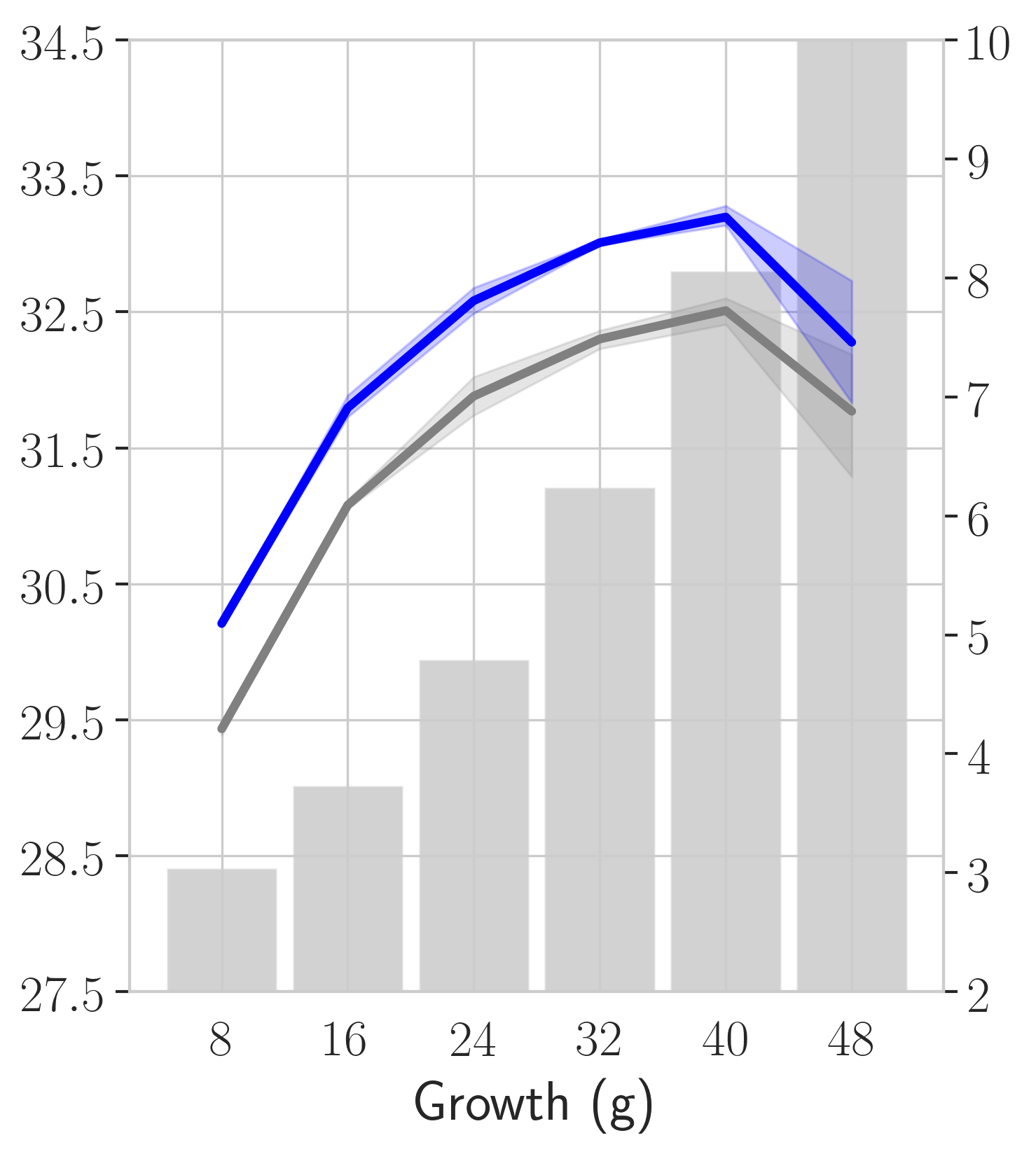}
\end{subfigure}
\begin{subfigure}{0.32\textwidth}
\caption{\label{fig:ablation:depth}$d=128,\; g=32$ } 
\includegraphics[width=\textwidth]{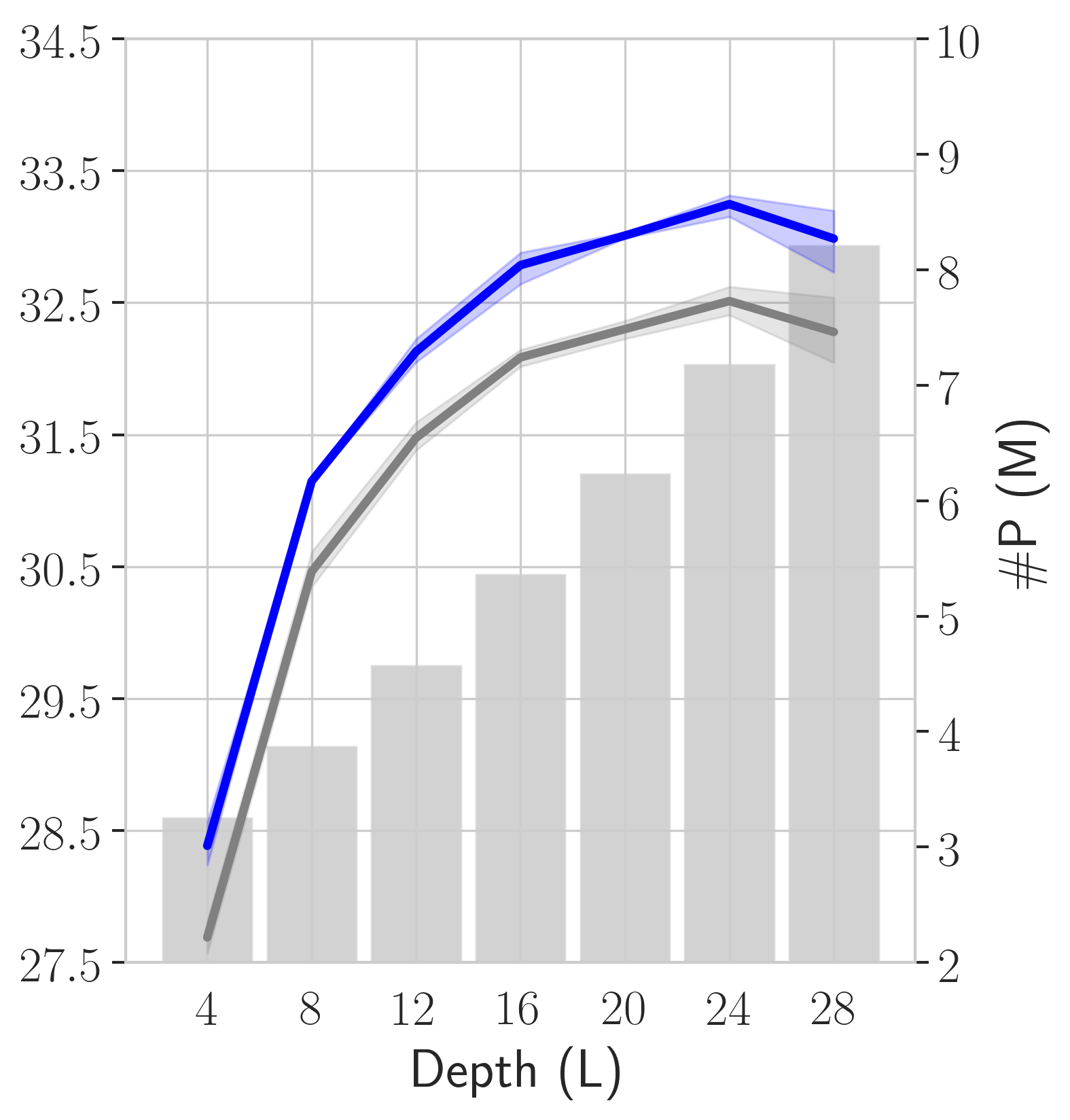} 
\end{subfigure}
\end{center}
\caption{The impact of token embedding size, number of layers ($L$), and growth rate ($g$) on the validation set BLEU scores. 
    In blue the results with beam search (width=5) and in gray with greedy decoding. The bars show the total number of parameters (in millions) for each setup.
}
\label{fig:ablation}
\end{figure*}

\paragraph{Baselines.}
For comparison with state-of-the-art architectures, we implemented a bidirectional LSTM encoder-decoder model with dot-product attention \cite{bahdanau15iclr, luong15emnlp} using PyTorch \cite{paszke17nipsw}, and used Facebook AI Research Sequence-to-Sequence Toolkit \cite{gehring17icml} to train the ConvS2S and  Transformer \cite{vaswani17nips} models on our data.

For the Bi-LSTM encoder-decoder, the encoder is a single layer bidirectional LSTM with input embeddings of size 128 and a hidden state of size 256 (128 in each direction). The decoder is a single layer LSTM with similar input size and a hidden size of 256, the target input embeddings are also used in the pre-softmax projection. For regularization, we apply a dropout of rate 0.2 to the inputs of both encoder and decoder and to the output of the decoder prior to softmax. As in \cite{bahdanau15iclr}, we  refer to this model as RNNsearch.

The ConvS2S model we trained has embeddings of dimension 256, a 16-layers encoder and 12-layers decoder. Each convolution uses $3\!\times\!1$ filters and is followed by a gated linear unit with a total of $2 \times 256$ channels. Residual connections link the input of a convolutional block to its output.
We first trained the default architecture for this dataset as suggested in FairSeq \cite{gehring17icml}, which has only 4 layers in the encoder and 3 in the decoder, but achieved better results with the deeper version described above.
The model is trained with label-smoothed cross-entropy ($\epsilon=0.1)$ using Nesterov accelerated gradient  with a momentum of 0.99 and an initial learning rate of 0.25 decaying by a factor of 0.1 every epoch. ConvS2S is also regularized with a dropout rate of 0.2.

For the transformer model,
we use token embeddings of dimension 512, and the encoder and decoder have 6 layers and 4 attention heads. For the inner layer in the per-position feed-forawrd network we use $d_{ff}=1024$.
We optimize the label-smoothed ($\epsilon=0.1)$ cross-entropy loss with Adam $(\beta_1=0.9, \beta_2=0.98, \epsilon=1e^{-8})$ \cite{kingma15iclr}. The learning rate starts from $1e^{-7}$ and is increased during 4,000 warm-up steps. Afterwards, the learning rate is set to $5e^{-4}$ and follows an inverse-square-root schedule \cite{vaswani17nips}. For the transformer we set the dropout to $0.3$.

\subsection{Experimental results}

\begin{table}
\begin{center}
{\small
\begin{tabular}{HHHHccHHHcc}
\toprule
              & embedding & growth  & depth & Model      & BLEU                & BLEU (best) & std(data) &  std  & Flops$\times10^5$ & \#params\\ 
\midrule
De-En         &    128    & 32      & 24    & Average    &  30.89 $ \pm$ 0.18  &    31.19    &           &  0.18 &     3.63    & 7.18M \\
De-En         &    128    & 32      & 24    & Max        &  33.25 $ \pm$ 0.1   &    33.32    &           &  0.1  &     3.44    & 7.18M \\ 
De-En         &    128    & 32      & 24    & Attn       &  31.55 $\pm$ 0.11   &    31.71    &           &  0.13 &     3.61    & 7.24M \\
De-En         &    128    & 32      & 24    & Max, gated &  32.99 $\pm$ 0.17   &    33.26    &           &  0.17 &     3.49    & 9.64M \\
De-En         &    128    & 32      & 24    & [Max, Attn]&  33.29 $\pm$ 0.14   &    33.47    &           &  0.14 &     3.51    & 7.24M \\
\bottomrule
\end{tabular}
}
\end{center}
\caption{BLEU scores of our model ($L\!=\!24, g\!=\!32, d_s\!=\!d_t\!=\!128$) on the validation set with different pooling operators and using gated convolutional units.}
\label{tab:ablation}
\end{table}

\paragraph{Architecture evaluation.}
In this section we explore the impact of several parameters of our model:  the  token embedding dimension, depth, growth rate and filter sizes. We also evaluate different aggregation mechanisms across the source dimension: max-pooling, average-pooling, and  attention.

In each chosen setting, we train five models with different initializations and report the mean and standard deviation of the validation set BLEU scores. We also state the number of parameters of each model and the computational cost of training, estimated in a similar way as  \citet{vaswani17nips}, based on the wall clock time of training and the GPU single precision specs.

In \tab{ablation} we see that using max-pooling instead average-pooling across the source dimension increases the performance with around 2.3 BLEU points. 
Scaling the average representation with $\sqrt{\lsrc}$ \Eq{avgpool} helped improving the performance but it is still largely outperformed by the max-pooling.
Adding gated linear units on top of each convolutional layer does not improve the BLEU scores, but increases the variance due to the additional parameters. 
Stand-alone self-attention \ie weighted average-pooling is slightly better than uniform average-pooling but it is still outperformed by max-pooling. 
Concatenating the max-pooled features (\Eq{maxpool}) with the representation obtained with self-attention (\Eq{attention}) leads to a small increase in performance, from 33.25 to 33.29. In the remainder of our experiments we only use max-pooling for simplicity, unless stated otherwise. 

In \fig{ablation} we consider the effect of the token embedding size, the growth rate of the network, and its depth. 
The token embedding size together with the growth rate $g$ control the dimension of the final feature used for estimating the emission probability. We generaly use the same embedding dimension for both languages i.e. $d=d_t=d_s$, thus the final representation is of size $f_L = 2 d + gL$.
In \fig{ablation} we see that a minimal dimension is required, in this case $d=128$, in order for the model to be complex enough and capture the training data statistics.
For embedding sizes between 128 and 256, the BLEU score slowly increases from 33 to 33.6

The depth of the network is of similar impact. Training deeper networks (from 4 to 24 layers)  increases the BLEU score by about 5 points.
An argument similar to the one about the growth rate can be made in this case too for networks with more than 24 layers.

The receptive field of our model is controlled by its depth and the filter size. In \tab{ablation:kernel}, we note that
narrower receptive fields are better than larger ones with less layers at equivalent complextities \eg comparing ($k\!=\!3, L\!=\!20$) to ($k\!=\!5, L\!=\!12$), and ($k\!=\!5, L\!=\!16$) with ($k\!=\!7, L\!=\!12$).

\begin{table}
\begin{center}
{\small
\begin{tabular}{Hccccc}
\toprule
           & $k$ & $L$ &  BLEU     & Flops$\times10^5$ & \#params\\ 
\midrule
De-En      &      3        &  16    &  32.40$\pm$0.08   &    2.47           &  4.32M \\ 
De-En      &      3        &  20    &  32.57$\pm$0.23   &    3.03           &  4.92M \\ 
\midrule
De-En      &      5        &   8    &  31.14$\pm$0.04   &    0.63           & 3.88M  \\ 
De-En      &      5        &  12    &  32.13$\pm$0.11   &    2.61           & 4.59M  \\ 
De-En      &      5        &  16    &  32.78$\pm$0.16   &    3.55           & 5.37M  \\ 
De-En      &      5        &  20    &  33.01$\pm$0.03   &    3.01           & 6.23M  \\ 
De-En      &      5        &  24    &  33.25$\pm$0.1   &    3.44           & 7.18M  \\ 
De-En      &      5        &  28    &  32.99$\pm$0.3   &    5.35           & 8.21M  \\ 
\midrule
De-En      &      7        &  12    &  31.81$\pm$0.2   &    2.76           & 5.76M  \\ 
De-En      &      7        &  16    &  32.43$\pm$0.36   &    3.13           & 6.94M  \\ 
\bottomrule
\end{tabular}
}
\end{center}
\caption{Performance of our model ($g\!=\!32, d_s\!=\!d_t\!=\!128$) for different 
filter sizes $k$ and depths $L$ and filter sizes $k$ on the validation set.}
\label{tab:ablation:kernel}
\end{table}

\paragraph{Comparison to the state of the art.}

\begin{table*}
\begin{center}
{\small
\begin{tabular}{cccc|cHc}
\toprule
\bf Word-based                                         & De-En     & \makecell{Flops\\ ($\times10^5$)} & \# prms  & En-De & \makecell{Flops\\ ($\times10^5$)} & \# prms \\
\midrule
{Conv-LSTM (MLE) \citep{bahdanau17iclr} }  & 27.56     &        & & & & \\ 
{Bi-GRU (MLE+SLE) \citep{bahdanau17iclr}}  & 28.53     &        & & & & \\ 
\midrule
{Conv-LSTM (deep+pos) \citep{gehring17acl} }& 30.4      &        & & & & \\ 
{NPMT + language model \cite{huang18iclr}}  & 30.08     &        & & 25.36 & & \\
\midrule
\midrule
\bf BPE-based                                          &           &        & & & & \\
\toprule
{ConvS2S (MLE+SLE) \cite{edunov18naacl}}   &  32.84    &       &       & & & \\ 
{Varational attention \cite{deng18arxiv}}  &  33.10    &       &        & & & \\
\midrule
{RNNsearch* \citep{bahdanau15iclr}}, V1 &  29.98    & 1.79  & 13M  & 25.04 &  &15M  \\
{ConvS2S** (MLE) \cite{gehring17icml}}, V1      &  32.31    &      1.35 & 21M   & 26.73 & & 22M\\ 

{Transformer** \citep{vaswani17nips}}, V1       &  34.42    &           & 46M   & \bf  28.23   &  & 48M \\ 
{Transformer** \citep{vaswani17nips}}, V2       &  \bf 34.44    &           & 52M   &  28.07   &  & 52M \\ 
\midrule
\midrule
\ours (this paper), V1                            &  33.86     &  & \bf 11M   & 27.21$  $  & &\bf 11M\\  
\ours (this paper), V2                            &  34.18    &  & 22M   & 27.99  & ~ & 22M\\

\bottomrule
\end{tabular}
}

\caption{Comparison to state-of-the art results on IWSLT German-English translation.
(*): results obtained using our implementation. (**): results obtained using FairSeq \cite{gehring17icml}.
\label{tab:sota}}
\end{center}
\end{table*}

We compare our results to the state of the art in \tab{sota} for both directions German-English (De-En) and English-German (En-De).

In this section, the parameters of our models are trained using label-smoothed cross-entropy ($\epsilon=0.1$) similarly to the \textit{ConvS2S} and \textit{Transformer} baselines. To successfuly train our models with large embeddings ($d=512$) we increase the dropout ($p=0.4$) and normalize the initial 2D grid. For decodig we use a beam-search of width 5 enhaced with length and coverage penalties \citep{wu16arxiv}.

Our model has about the same number of parameters as RNNsearch (with V1 vocbaularies), yet improves  performance by 3.88 BLEU points.
It is also  better than the recent work of \citet{deng18arxiv} on recurrent architectures with variational attention.

Our model outperforms its 1D convolutional counterpart \citet{gehring17icml} in both translation directions and is competitive with transformer (0.3 points behind) while having about 2 to 4 times fewer parameters.

\begin{figure}
    \includegraphics[width=\columnwidth]{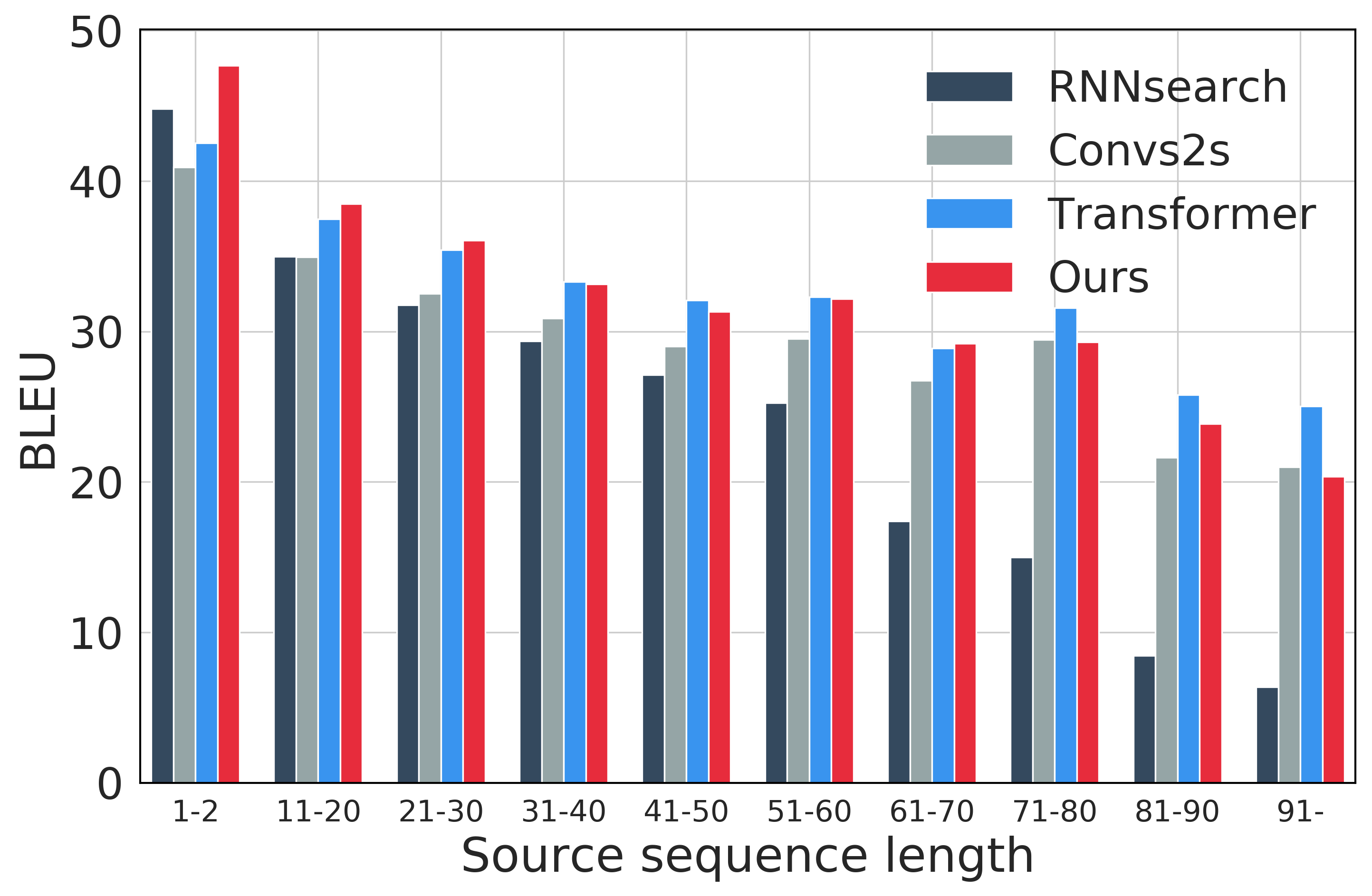}
    \caption{BLEU scores across sentence lengths.}
    \label{fig:length}
\end{figure}

\paragraph{Performance across sequence lengths.}
In \fig{length} we consider translation quality as a function of 
sentence length, and compare our model to RNNsearch, ConvS2S and Transformer.
Our model gives the best results across all sentence lengths, except for the longest ones where ConvS2S and Transformer are better.
Overall, our model combines the strong performance of RNNsearch on short sentences with good performance of ConvS2S and Transformer on longer ones.

\begin{figure*}
\begin{center}
\begin{subfigure}{.49\textwidth}
\includegraphics[height=.82\linewidth]{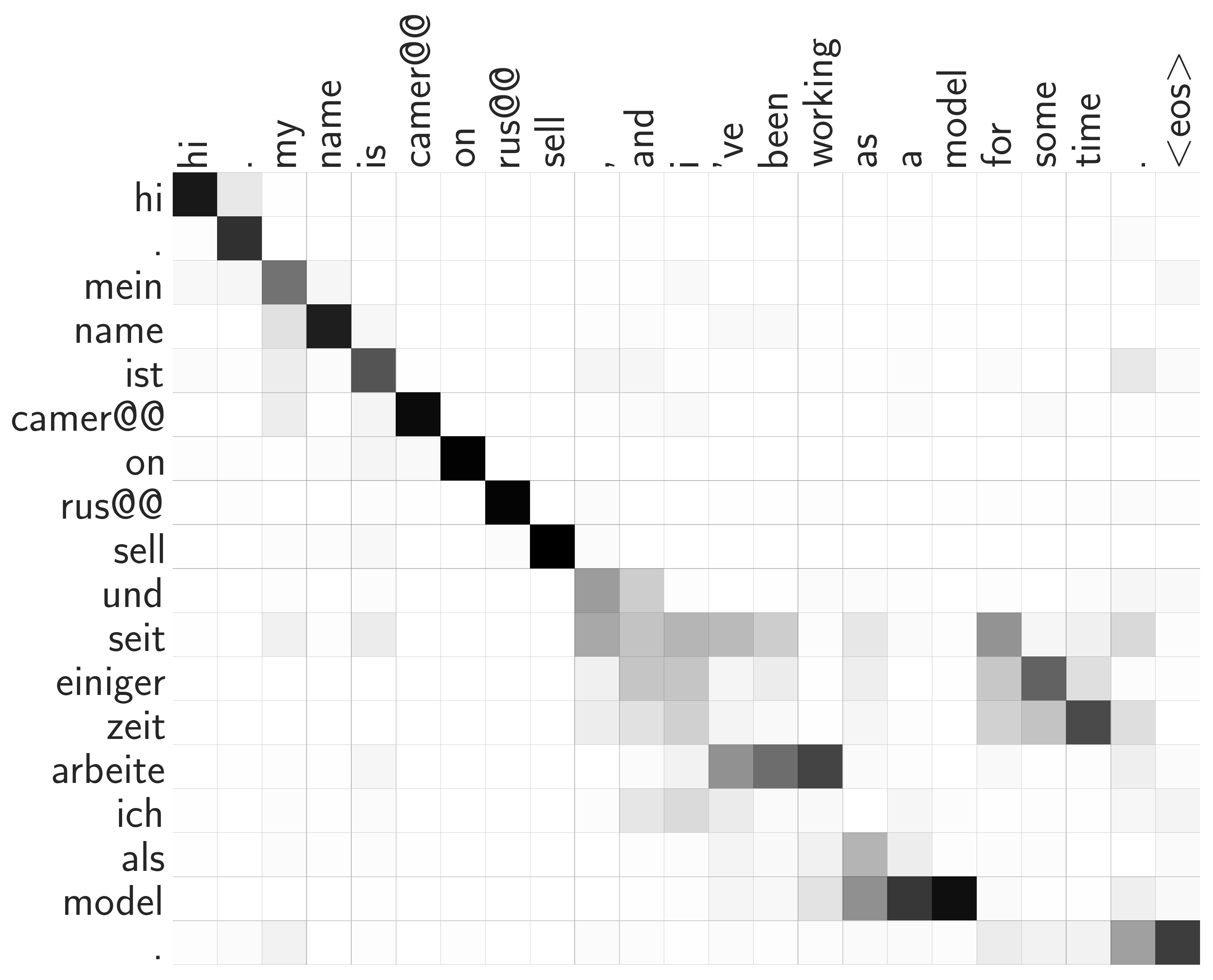}
\caption{Max-pooling}
\end{subfigure}
\hfill
\begin{subfigure}{.49\textwidth}
\includegraphics[height=.82\linewidth]{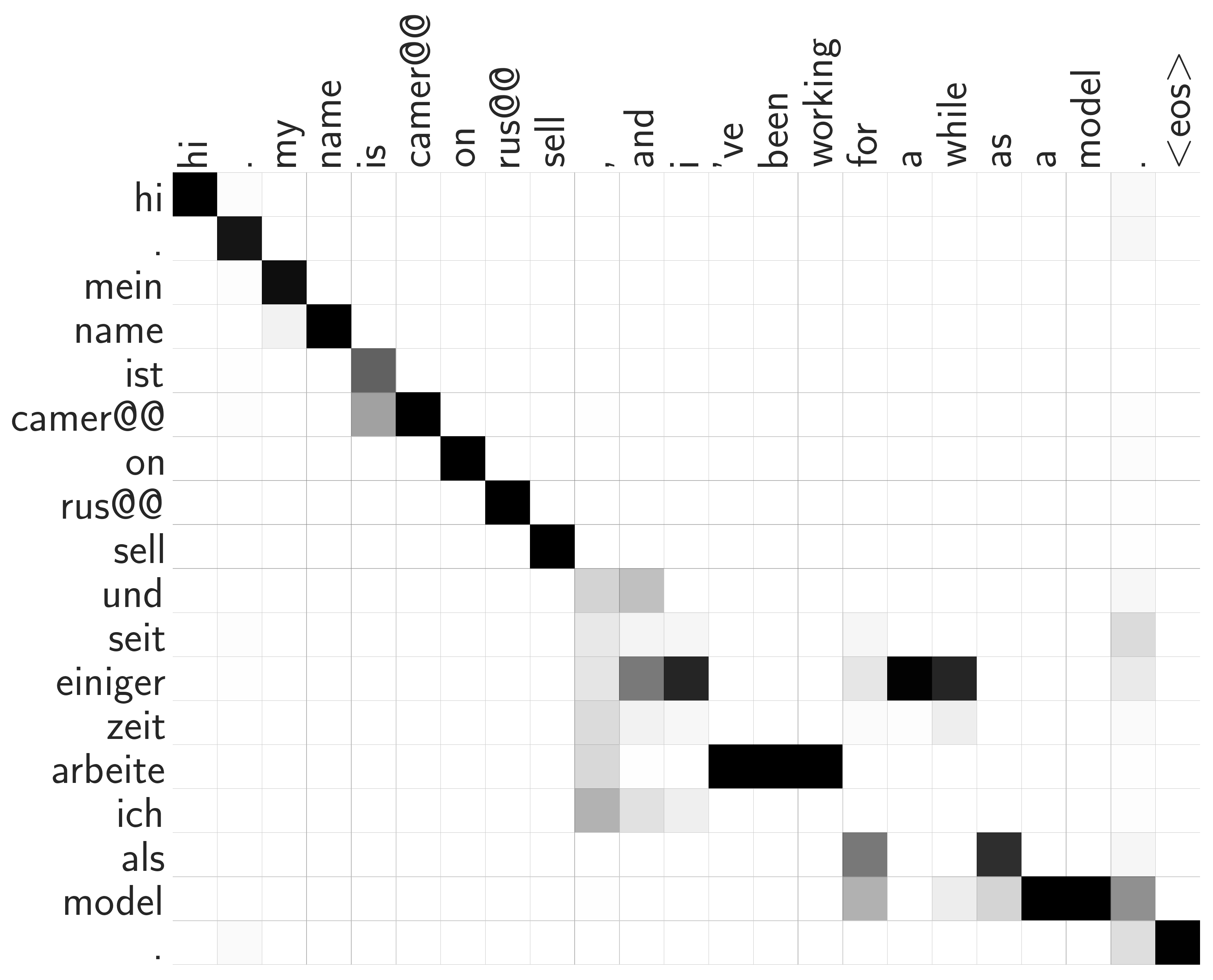}
\caption{Self-attention}
\end{subfigure}
\vspace{5mm}
\begin{subfigure}{.49\textwidth}
\includegraphics[height=.94\linewidth]{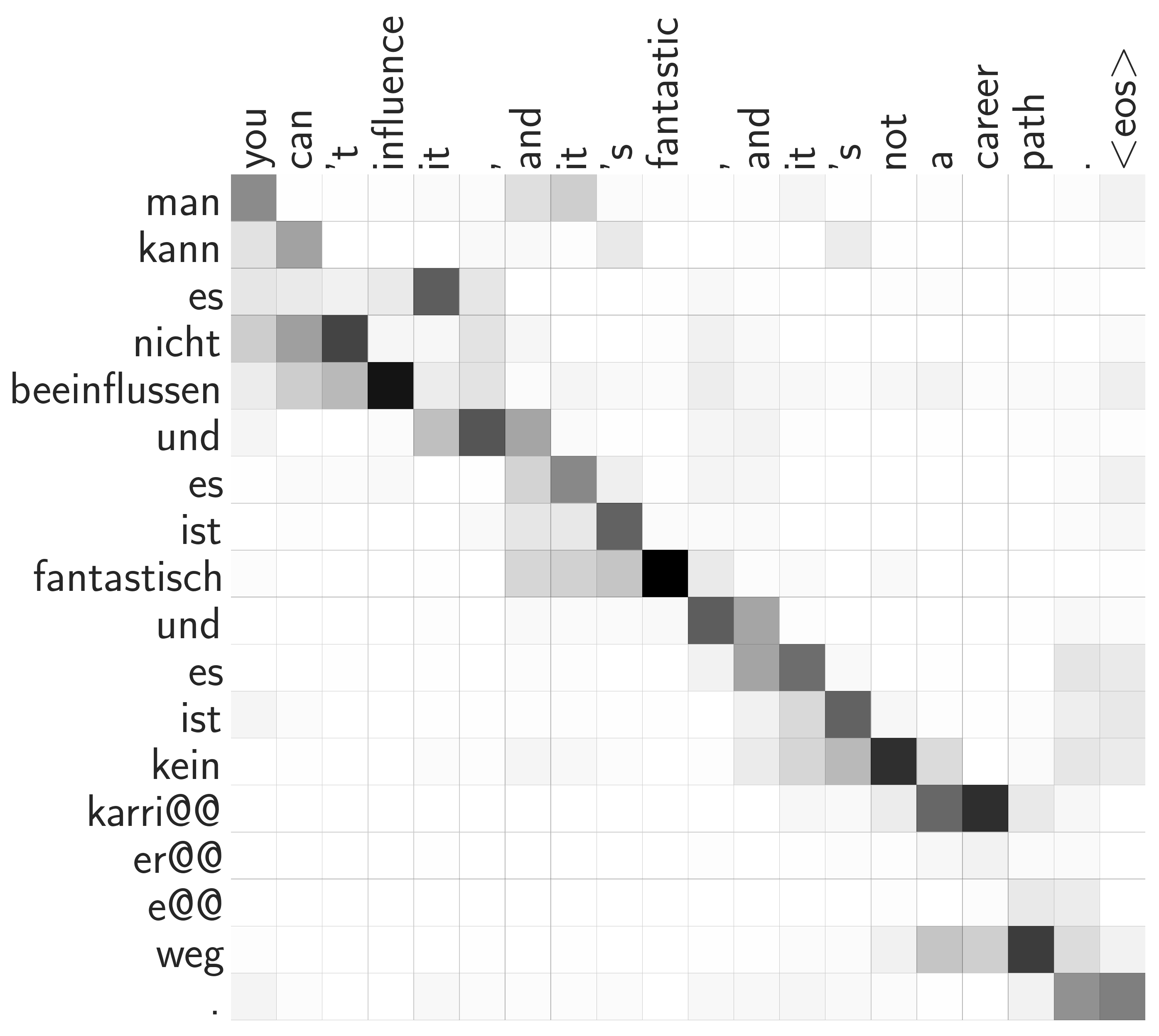}
\caption{Max-pooling}
\end{subfigure}
\hfill
\begin{subfigure}{.49\textwidth}
\includegraphics[height=.94\linewidth]{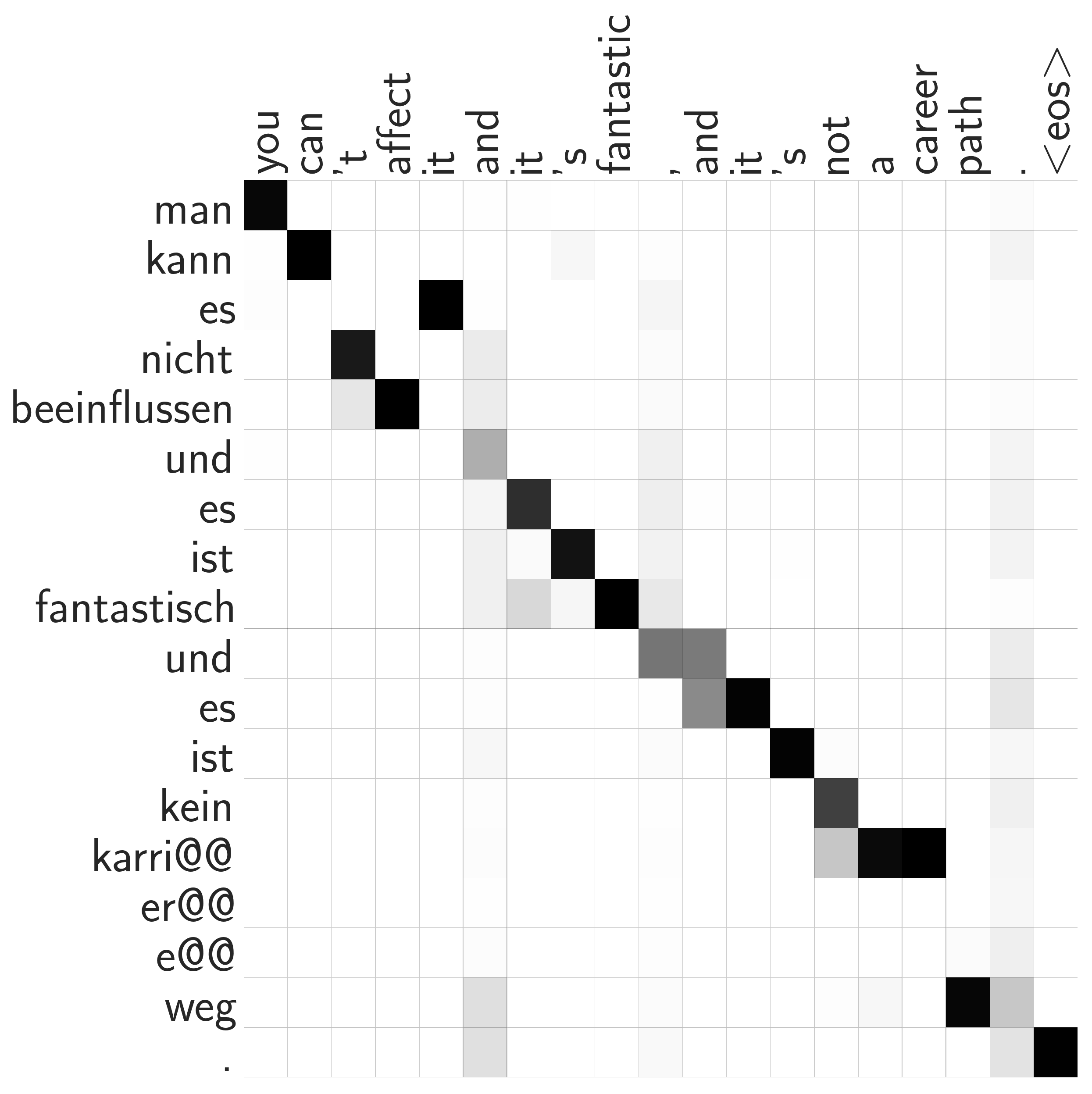}
\caption{Self-attention}
\end{subfigure}
\vspace{5mm}
\begin{subfigure}{.49\textwidth}
\includegraphics[height=.65\linewidth]{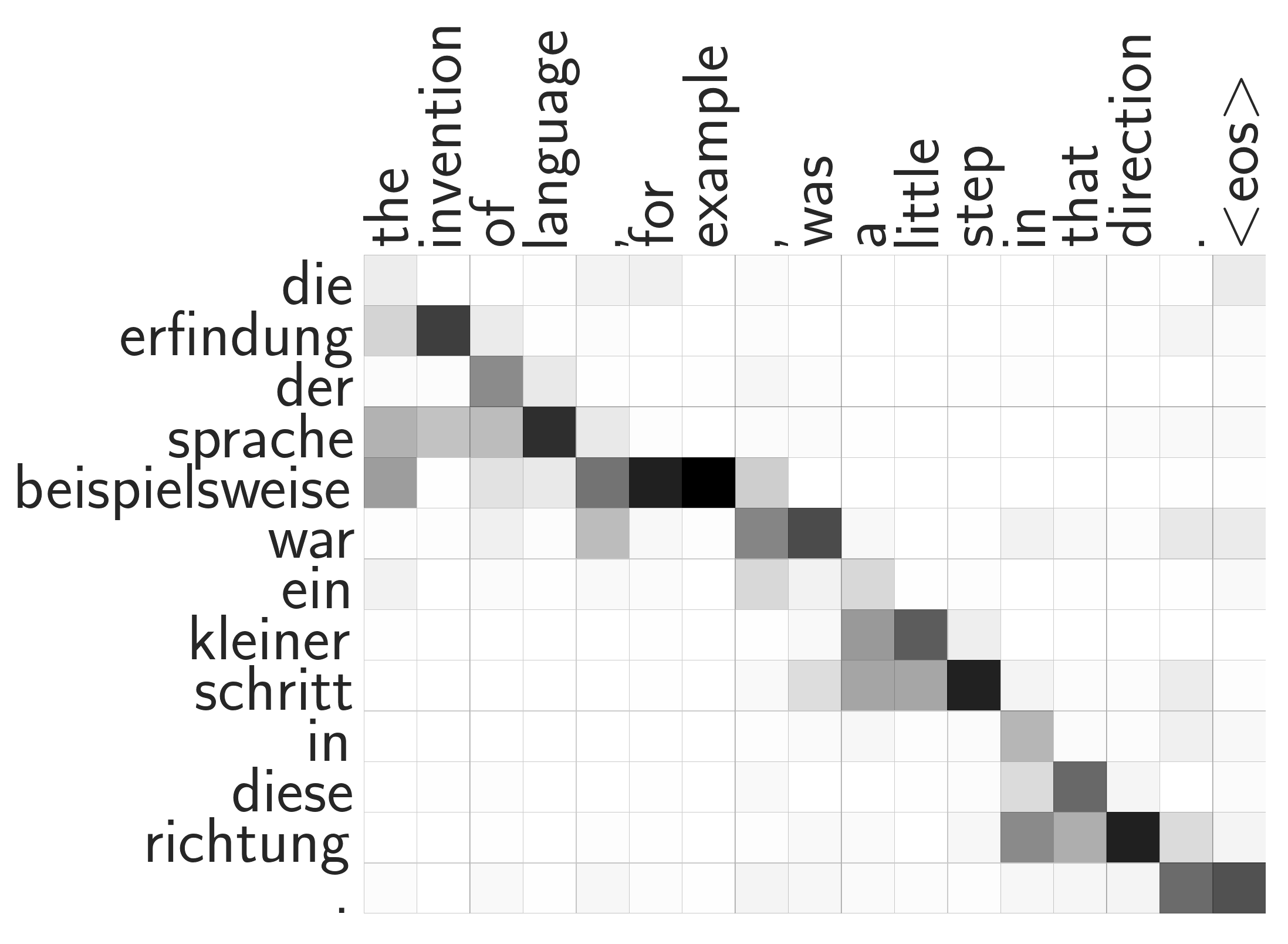}
\caption{Max-pooling}
\end{subfigure}
\hfill
\begin{subfigure}{.49\textwidth}
\includegraphics[height=.65\linewidth]{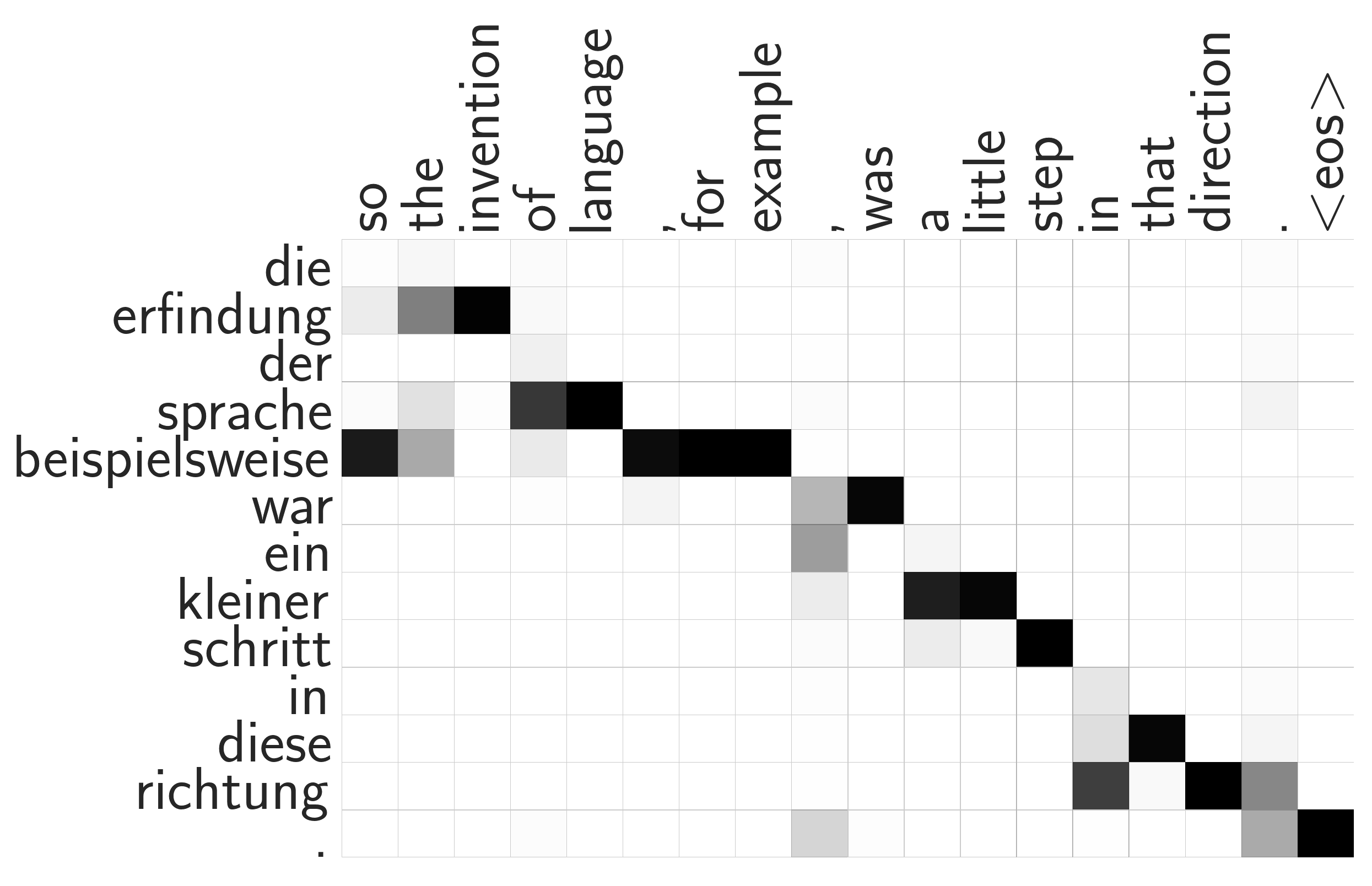}
\caption{Self-attention}
\end{subfigure}
\end{center}
\caption{Implicit BPE token-level alignments produced by our \ours model. For the max-pooling aggregation we visualize $\alpha$ obtained with \Eq{implicit} and for self-attention the weights $\rho$ of \Eq{attention1}.
}
\label{fig:attention-implicit}
\end{figure*}

\paragraph{Implicit sentence alignments.}
Following the method described in Section \ref{sec:model}, we illustrate in \fig{attention-implicit} the implicit sentence alignments the max-pooling operator produces in our model. For reference we also show the alignment produced by our model using self-attention.
We see that with both max-pooling and attention qualitatively similar 
implicit sentence alignments emerge. 

Notice in the first example how the max-pool model, when writing \emph{I've been working}, looks at \emph{arbeite} but also at \emph{seit} which indicates the past tense of the former.
Also notice some cases of non-monotonic alignment. In the first example  \emph{for some time}  occurs at the end of the English sentence, but \emph{seit einiger zeit} appears earlier in the German source. 
For the second example there is non-monotonic alignment around the negation at the start of the sentence.
The first example illustrates the ability of the model to translate proper names by breaking them down into BPE units.
In the second example the German word \emph{Karriereweg} is broken into the four BPE units \emph{karri,er,e,weg}. The first and the fourth are mainly used to produce the English \emph{a carreer}, while for the subsequent \emph{path} the model looks at \emph{weg}.

Finally, we can observe an interesting pattern in the alignment map for several phrases across the three examples. A rough lower triangular pattern is observed for the English phrases \emph{for some time}, \emph{and it's fantastic}, \emph{and it's not}, \emph{a little step}, and \emph{in that direction}. 
In all these cases the phrase seems to be decoded as a unit, where features are first taken across the entire corresponding source phrase, and progressively from the part of the source phrase that remains to be decoded.

\section{Conclusion}
\label{sec:conclusion}

We presented a novel neural machine translation architecture that departs from the encoder-decoder paradigm. 
Our model jointly encodes the source and target sequence into a deep feature hierarchy in which the source tokens are embedded in the context of a partial target sequence. 
Max-pooling over this joint-encoding along the source dimension is used to map the features to a prediction for the next target token. 
The model is implemented as 2D CNN based on DenseNet, 
with masked convolutions to ensure a proper autoregressive factorization of the conditional probabilities.

Since each layer  of our model re-encodes the input tokens in the context of the target sequence generated so far, the model has attention-like properties in every layer of the network by construction.
Adding an explicit self-attention module therefore has a very limited, but positive, effect. 
Nevertheless, the max-pooling  operator in our model generates implicit sentence alignments that are qualitatively similar to the ones generated by attention mechanisms.
We evaluate our model on  the IWSLT'14 dataset, translation German to English and vice-versa. 
We obtain excellent BLEU scores that compare favorably with the state of the art, while using a conceptually simpler model with fewer parameters.

We hope that our alternative joint source-target encoding sparks interest in other alternatives to the encoder-decoder model.
In the future, we plan to explore hybrid approaches in which the input to our joint encoding model is not provided by token-embedding vectors, but the output of 1D source and target embedding networks, \eg \mbox{(bi-)LSTM} or 1D convolutional.
We also want to explore how our model can be used to translate across multiple language pairs.

Our PyTorch-based implementation is available at \url{https://github.com/elbayadm/attn2d}.
\\\\
{{\bf Acknowledgment.} 
This work has been partially supported by the grant ANR-16-CE23-0006 ``Deep in France'' and ANR-11-LABX-0025-01 ``LabEx PERSYVAL''.
} {}

\bibliographystyle{acl_natbib_nourl}
\bibliography{jjv}

\end{document}